\documentclass[]{spie}  

\usepackage{amsmath,amsfonts,amssymb}
\usepackage{graphicx}
\usepackage[colorlinks=true, allcolors=blue]{hyperref}
\usepackage{makecell}
\usepackage{subcaption} 
\usepackage{dirtytalk} 
\usepackage{floatrow}
\usepackage{enumitem}
\newfloatcommand{capbtabbox}{table}[][\FBwidth]

\title{Synthetic training data generation for deep learning based quality inspection}

\author[1]{Pierre Gutierrez}
\author[2]{Maria Luschkova}
\author[1]{Antoine Cordier}
\author[1]{Mustafa Shukor}
\author[2]{Mona Schappert}
\author[2]{Tim Dahmen}

\affil[1]{Scortex, 22 rue Berbier du Mets, Paris, France}
\affil[2]{German Research Center for Artificial Intelligence (DFKI), Saarbrücken, Germany}

\authorinfo{For further author information, send correspondence to P. Gutierrez\\E-mail: pgutierrez@scortex.io}

\begin{document} 
\maketitle



\begin{abstract}
Deep learning is now the gold standard in computer vision-based quality inspection systems. In order to detect defects, supervised learning is often utilized, but necessitates a large amount of annotated images, which can be costly: collecting, cleaning, and annotating the data is tedious and limits the speed at which a system can be deployed as everything the system must detect needs to be observed first. This can impede the inspection of rare defects, since very few samples can be collected by the manufacturer.
In this work, we focus on simulations to solve this issue. We first present a generic simulation pipeline to render images of defective or healthy (non defective) parts. As metallic parts can be highly textured with small defects like holes, we design a texture scanning and generation method. We assess the quality of the generated images by training deep learning networks and by testing them on real data from a manufacturer. We demonstrate that we can achieve encouraging results on real defect detection using purely simulated data. Additionally, we are able to improve global performances by concatenating simulated and real data, showing that simulations can complement real images to boost performances. Lastly, using domain adaptation techniques helps improving slightly our  final results.

\end{abstract}

\section{INTRODUCTION}
\label{sec:intro}  
The main fundamental challenge of applying deep supervised learning to quality inspection in manufacturing processes is the availability of training data. In the case of cast products, there is a broad spectrum of potential abnormalities like scratches, cracks,  blow holes, or chipping to be detected by automatized inspection systems. However, due to the rare occurrence of such defective parts, it is almost impossible to gather training data exhibiting defects in all possible forms (i.e. with maximum variability). This inevitably leads to class imbalance during training. Furthermore, annotating available data usually constitutes a tremendous effort, often even necessitating expert knowledge. A promising approach to overcome the data availability problem is to use a graphic simulator which automatically generates labeled data \cite{tremblay2018training}. Indeed, models can theoretically learn more from the photo-realistic simulacrum than from the real environment: it is possible to simulate rare (or even unseen) events, as well as generate widely distributed variations of these events in a data set. This helps to train deep learning algorithms more effectively, leading to more robust and adaptive networks.         

In this paper, we evaluate the possibility to generate images of large enough quality and quantity for learning deep models for visual inspection purposes. Our plan is as follow: in section \ref{sec:related_work} we review the literature. We then detail our methodology (simulation pipeline and deep learning approach) in section \ref{sec:method}. Our results are presented in section \ref{sec:experiments}. We demonstrate that it is possible to improve detection capabilities by adding simulated images, but that using simulated data alone is not enough to get industrial grade performances.

\section{Related work}
\label{sec:related_work}

Quality control using \textbf{deep learning computer vision techniques} usually relies on supervised learning \cite{dong2019pga}. To this end, convolutional neural networks (CNNs) are trained to classify and localize defects on images: this can typically be achieved either using semantic segmentation networks such as U-Net \cite{ronneberger2015u} and DeepLab \cite{chen2017deeplab}, or object detection architectures like YOLO\cite{redmon2016you} or RetinaNet \cite{lin2017focal}. However, defect occurrence can be a rare event. Thus, collecting enough data to properly learn to detect these defects may not always be possible. In such cases, practitioners can rely on unsupervised anomaly detection \cite{bergmann2019mvtec} or make use of simulated data \cite{zambal2019end}, which is the focus of our paper.

The use of \textbf{simulated data in deep learning} has been widely studied in the field of autonomous driving \cite{li2019aads}, robotics \cite{tobin2017domain}, and more broadly for object detection on images \cite{tremblay2018training}. Simulations are frequently utilized for tasks where human labeling is either hard or costly - such as pose estimation \cite{chen2016synthesizing,sundermeyer2018learning}, depth estimation \cite{langlois2019fly}, or deep learning-based rendering \cite{deschaintre2018single} - in order to lower the cost of human annotation or to learn theoretical cases for which no real data exist. Models trained on simulated data may not generalize well to real cases: this is commonly known as the \textit{domain gap} problem. Three techniques help solving this issue: improving photorealism (making the synthetic input domain closer to the real one), synthetic domain randomization (randomizing the synthetic input domain enough to cover the real domain), and domain adaptation (train the model in a way that the domains align with one another).

\textbf{Improved photorealism} is essential for learning strong detectors \cite{movshovitz2016useful}. It has recently been shown that rendering 3D models in physically realistic positions can have a significant impact on performances\cite{hodavn2019photorealistic}, compared to randomly rendering the same models in random positions on random backgrounds\cite{dwibedi2017cut}. However, in order to generate simulated data diverse enough for learning, photorealism has to be coupled with domain randomization \cite{tobin2017domain,prakash2019structured}.
The idea behind \textbf{domain randomization} \cite{tobin2017domain} was initially to randomize the simulation process so that the real domain could become a subset of the simulated one. This has successfully been applied to autonomous driving \cite{tremblay2018training} and industrial applications \cite{lee2019automatic}. The concept has now been extended to “automated domain adaptation” \cite{akkaya2019solving} (in which the randomization increases over time in a curriculum learning fashion) and to “structured domain randomization” \cite{prakash2019structured} (for which domain randomization does not come at the cost of photorealism). 
In general, \textbf{domain adaptation} grant benefits when dealing with data from at least two distinct distributions or domains. This is the case when training on simulated data while trying to generalize on real data. The common paradigm of domain adaptation states that while the source (simulation) domain labels are available, the target (real) domain labels are not. For a more complete overview of unsupervised domain adaptation, we refer the reader to the review by Toldo, Marco, et al. \cite{toldo2020unsupervised}. Note that it is also possible to leverage supervision in the target domain when labels are available (ASS \cite{wang2020alleviating}). Overall, three different types of domain adaptation can be distinguished: input space, feature space, and output space domain adaptations. In input space adaptation, images from the two domains are aligned so that they get similar image statistics. This can be done using GANs \cite{goodfellow2014generative, hoffman2018cycada, dundar2018domain} or Fourier transforms\cite{yang2020fda}. In feature space adaptation, a regularization term is used to penalize unaligned features. A very common way to do this is to use an adversarial loss (DANN \cite{ganin2016domain}, FCNs \cite{hoffman2016fcns}, Wasserstein DANN \cite{shen2018wasserstein}), though more traditional methods also exist: notably, CORAL\cite{sun2016deep} makes use of moments. Finally, output space adaptation can be used to align the network predictions between the source and the target.\cite{vu2019advent}.

To generate the synthetic training data per se, we follow the concept of \textbf{partial modeling} \cite{dahmen2019digital}. A partial (or shallow) model of reality can be understood as the replication of a subset of specific features from a real-life scenario. For example, in order to synthesize images of a cast part under the partial modeling hypothesis, one can concentrate its efforts on modeling the shape, texture, and orientation of the part, as well as on tuning lighting and camera parameters, rather than focusing on a perfect modeling of the highly complex properties of the casting material.

\newlength{\twosubht}
\newsavebox{\twosubbox}

\begin{figure}
    \sbox\twosubbox{
      \resizebox{\dimexpr.9\textwidth-1em}{!}{
        \includegraphics[height=3cm]{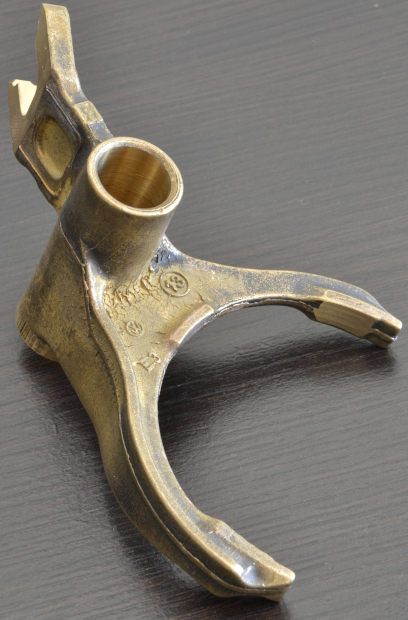}
        \includegraphics[height=3cm]{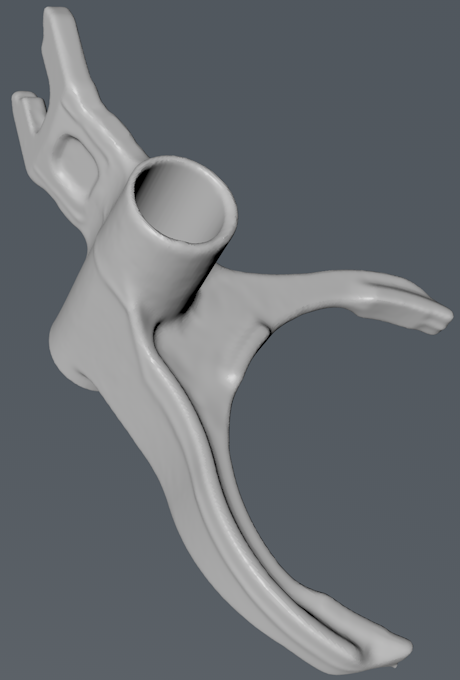}
        \includegraphics[height=3cm]{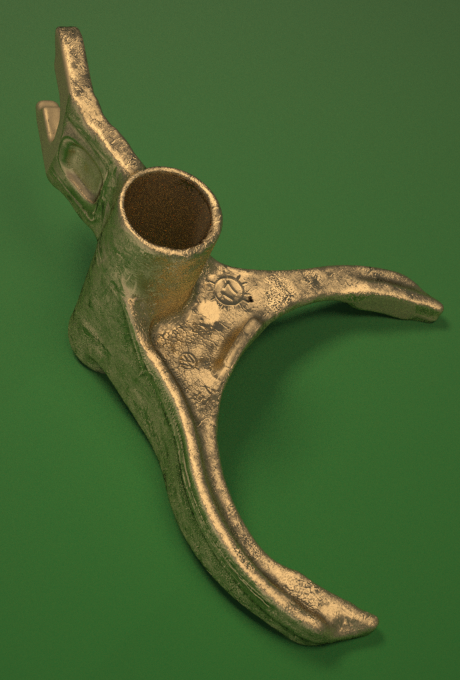}
        \includegraphics[height=3cm]{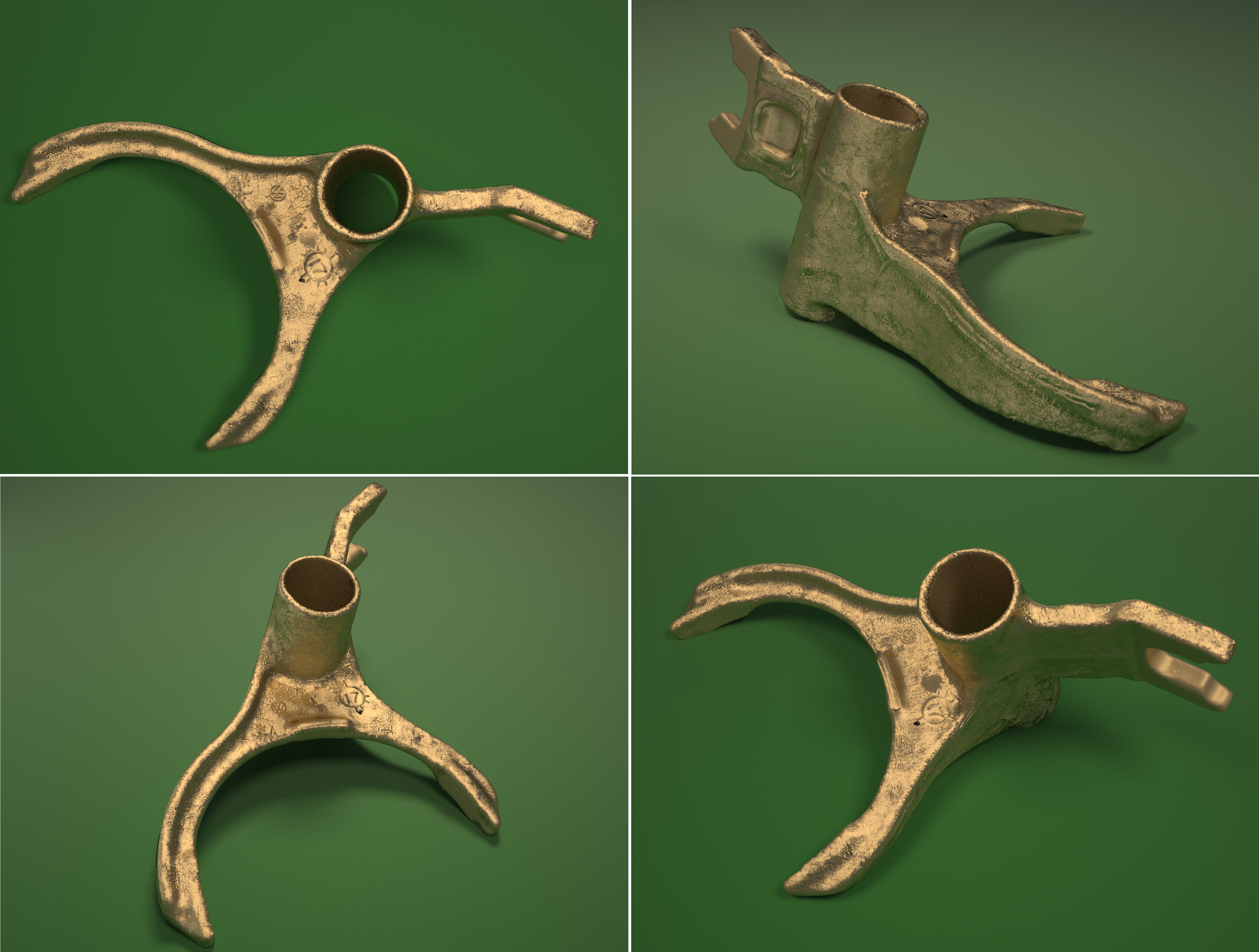}
      }
    }
    \setlength{\twosubht}{\ht\twosubbox}
    \centering
    \subcaptionbox{\label{real_VW}}{
      \includegraphics[height=\twosubht]{real_VW.JPG}
    }\quad
    \subcaptionbox{\label{VW_model_cropped}}{
      \includegraphics[height=\twosubht]{VW_model_cropped.png}
    }\quad
    \subcaptionbox{\label{VW_rendered_cropped}}{
      \includegraphics[height=\twosubht]{VW_rendered_v2.png}
    }\quad
    \subcaptionbox{\label{rendering_examples}}{
      \includegraphics[height=\twosubht]{VW_renderings.png}
    }
    \caption{(a) Photograph of the Volkswagen gear fork. (b) A 3D model, reconstructed photogrammetrically. (c) Synthesized image of Volkswagen gear fork with a hole defect (image center). (d) Example renderings of Volkswagen gear fork.}
    \label{VW}
\end{figure}

\section{MATERIALS AND METHODS}
\label{sec:method}  

The goal of this paper is to create a general pipeline for synthesizing data of extensive variance for manufacturing processes. It does so by bringing together texture scanning, texture synthesizing, parametric modeling of defects, and rendering. Due to the confidentiality of the manufacturer parts, we are unable to share any real or simulated images of our client's cast products. Therefore, in the following sections, we illustrate our image simulation pipeline on a purchased Volkswagen gear fork (Figure \ref{VW}).

\subsection{Geometry capture}
In our simulation pipeline, we separate the geometry capturing process into a coarse-grained 3D model capture of the object and a fine-grained texture scan. Usually, manufacturers have 3D models of their cast products. Otherwise, depending on the geometrical complexity of the object, a corresponding 3D model can be either created by hand or generated automatically by 3D laser scanners. In our showcase simulations, however, we capture the 3D model of a Volkswagen gear fork using photogrammetry \cite{remondino2005photogrammetry} (Figure \ref{VW_model_cropped}).    

\begin{figure}
    \sbox\twosubbox{
      \resizebox{\dimexpr.9\textwidth-1em}{!}{
        \includegraphics[height=3cm]{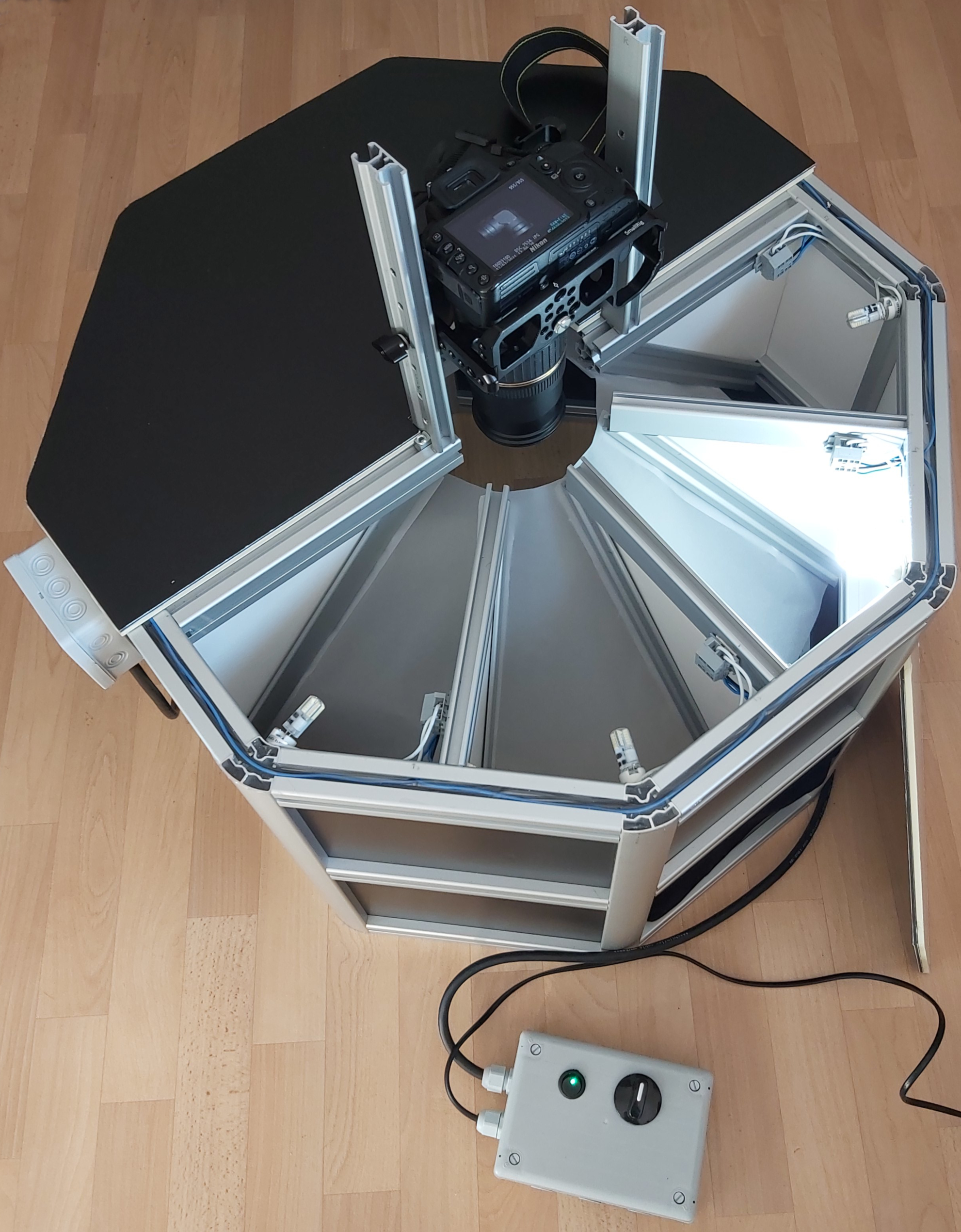}
        \includegraphics[height=3cm]{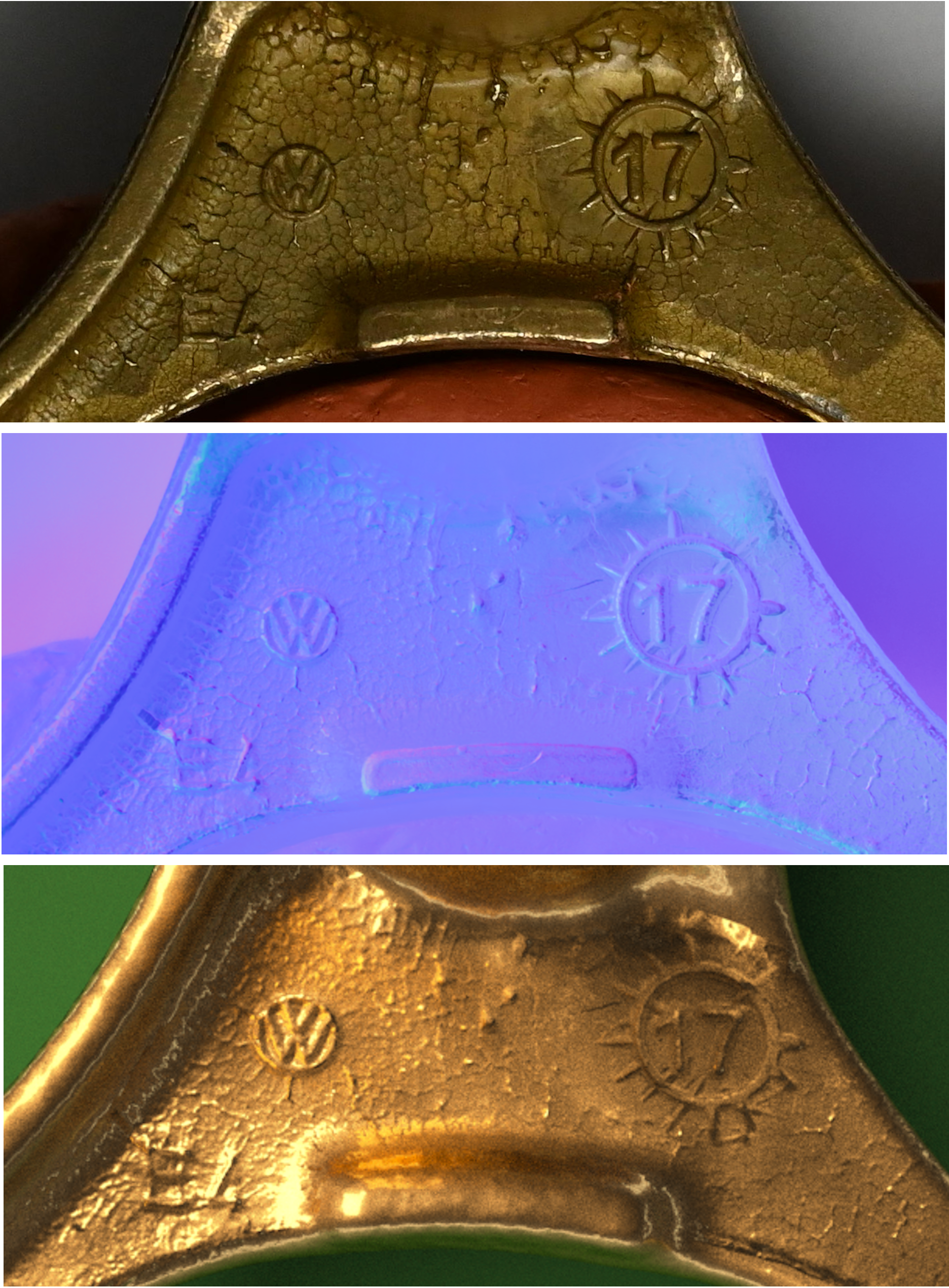}
        \includegraphics[height=3cm]{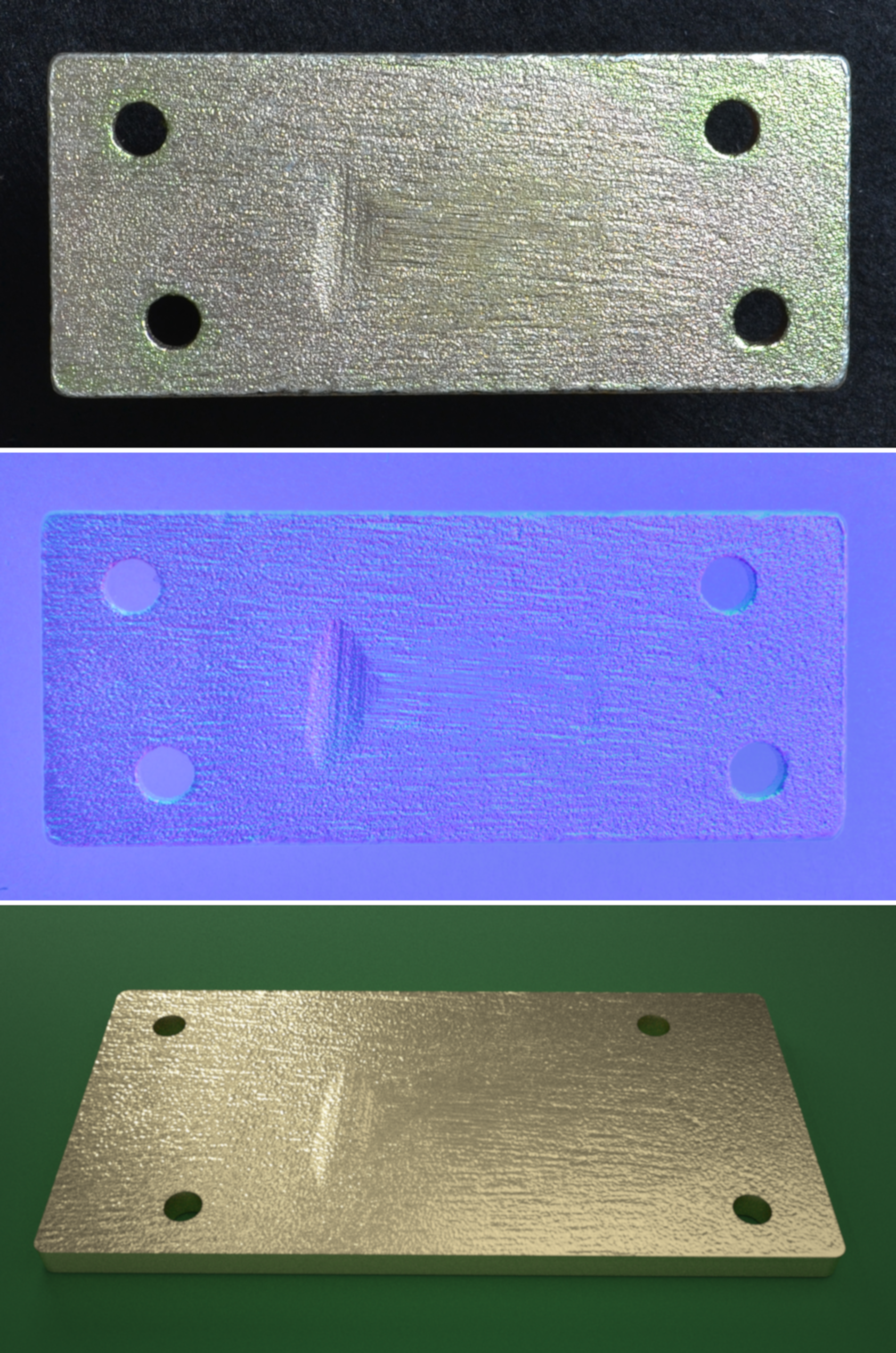}
        \includegraphics[height=3cm]{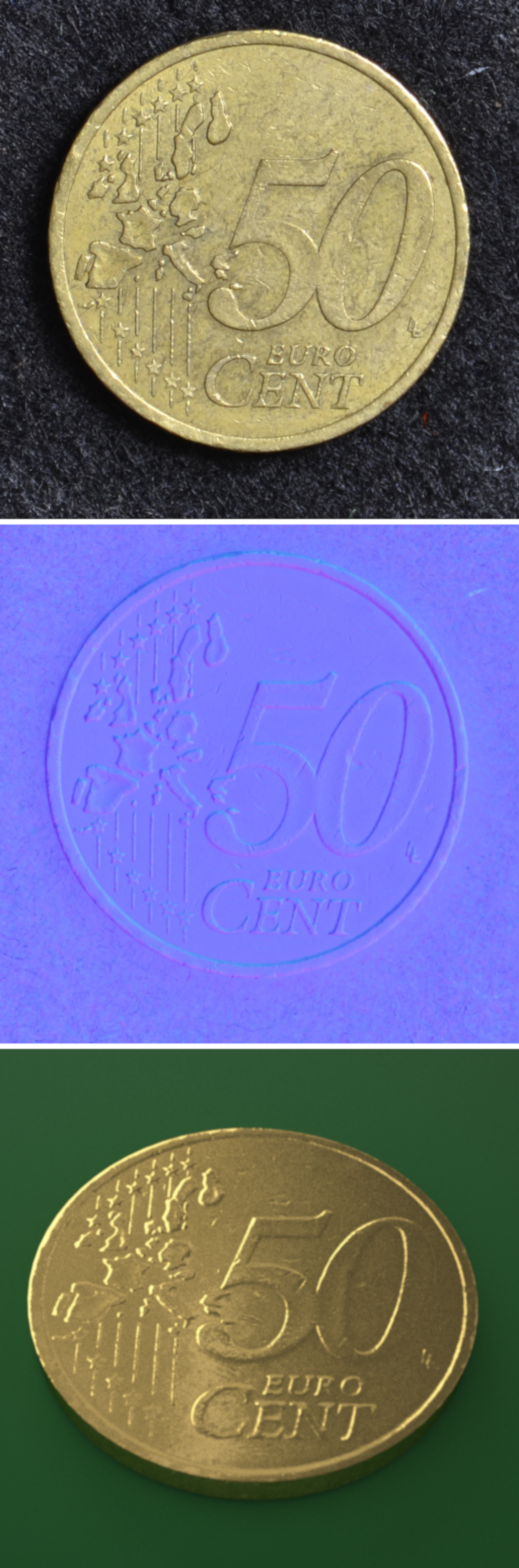}
      }
    }
    \setlength{\twosubht}{\ht\twosubbox}
    \centering
    \subcaptionbox{\label{scan_box}}{
      \includegraphics[height=\twosubht]{scan_box.jpg}
    }\quad
    \subcaptionbox{\label{textures_VW}}{
      \includegraphics[height=\twosubht]{textures_VW.png}
    }\quad
    \subcaptionbox{\label{textures_cast_part}}{
      \includegraphics[height=\twosubht]{textures_cast_part.png}
    }\quad
    \subcaptionbox{\label{textures_50_cents}}{
      \includegraphics[height=\twosubht]{textures_50_cents.png}
    }
    \caption{Our experimental photometric stereo-based setup with three target objects. (a) Scanning box. (b)-(d) First row: photographs of three target objects. Second row: reconstructed normal maps. Third row: textured 3D models. Renderings do not use exact color and reflectance information in order to focus on geometric aspects.}
\end{figure}

To capture fine-grained texture details, we use the photometric stereo technique from Woodham et al. \cite{woodham1980photometric}. This well-established method reconstructs 3D shapes from 2D source images, which are taken from the same viewpoint, but under different illumination directions. Consider the experimental setup shown in Figure \ref{scan_box}. A camera and eight light sources constitute the photometric stereo system: the camera is fixed at the center of the system, surrounded by eight light-emitting diode (LED) light sources arranged in an octagon. The slant angle of each light source is set to 45 degrees. The system is covered with foam boards to block any external light. To capture a flat texture of a sample, we place the sample under the camera and illuminate it with the different light sources in turn while taking perfectly aligned photos. To compute surface normal maps from the captured image stacks, we use the photometric stereo library available in Substance Designer tool (Figure \ref{textures_VW}-\ref{textures_50_cents}, second row). The reconstructed normal maps are used as bump maps \cite{blinn1978simulation} to texture the 3D models (Figure \ref{textures_VW}-\ref{textures_50_cents}, third row). 

\subsection{Texture randomization}
One limitation of textures scanned with the aforementioned method is their non-stochastic nature: the scan remains a static image of the material in a given condition. To synthesize new textures of typical casting surface finish, we apply a dictionary-based approach of exemplar-based inpainting \cite{criminisi2004inpainting,trampert2018inpainting}. This technique is especially well-suited for situations where only few texture scans are available. First, we build a texture dictionary from scanned normal samples (Figure \ref{inpainting_dictionary}). Next, we manually choose small texture patches which were not included in the dictionary: they are called seed points (Figure \ref{texturing_examples}, first column). Starting with the seed point texture, the inpainting algorithm iteratively fills empty regions around the patch by inserting the nearest neighbor patches from the dictionary. Since every following patch is inserted at the seed boundaries, the resulting synthesized textures exhibit consistent and coherent content (Figure \ref{texturing_examples}, second column).

\subsection{Defect generation}
\begin{figure}
    \sbox\twosubbox{
      \resizebox{\dimexpr.9\textwidth-1em}{!}{
        \includegraphics[height=3cm]{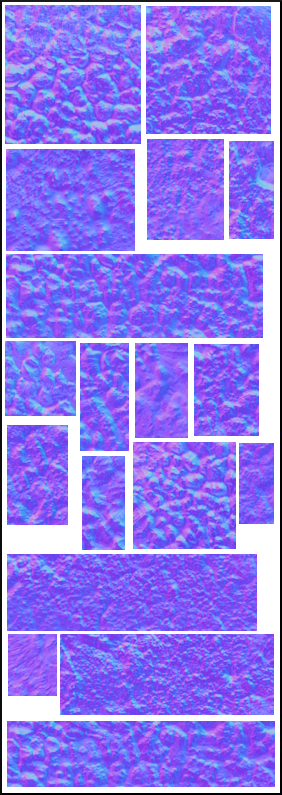}
        \includegraphics[height=3cm]{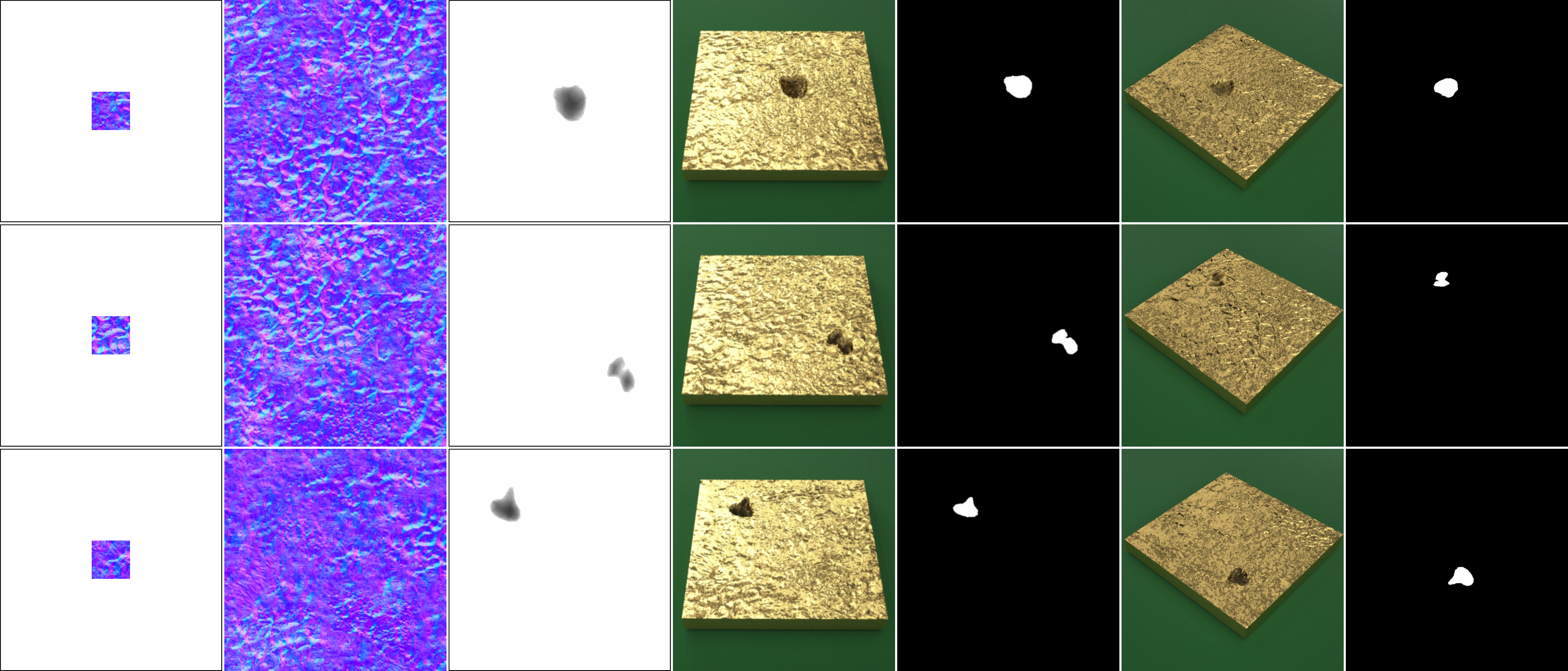}
        \includegraphics[height=3cm]{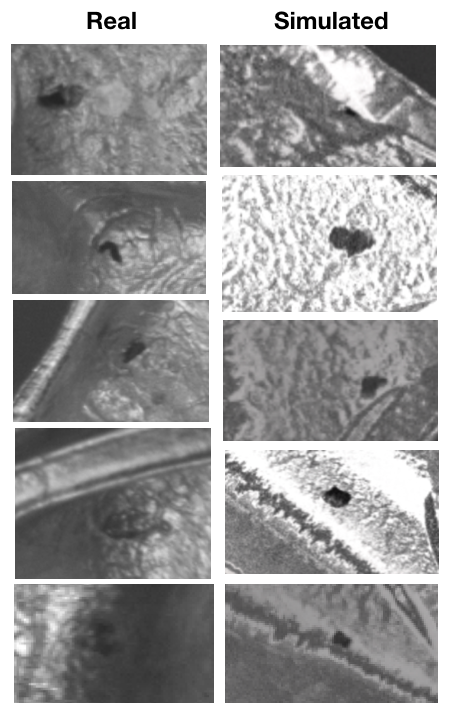}
      }
    }
    \setlength{\twosubht}{\ht\twosubbox}
    \centering
    \subcaptionbox{\label{inpainting_dictionary}}{
      \includegraphics[height=\twosubht]{inpainting_dictionary.png}
    }\quad
    \subcaptionbox{\label{texturing_examples}}{
      \includegraphics[height=\twosubht]{texturing_examples_v2.png}
    }\quad
    \subcaptionbox{\label{vertical}}{
      \includegraphics[height=4.75cm]{vertical.png} 
    }
    \caption{Texture and defect generation. (a) Excerpt of the texture dictionary. (b) Three examples of texture and defect generation. First column: seed patches for texture inpainting. Second column: synthesized normal maps. Third column: synthetically generated defects. Fourth and sixth columns: rendered 3D models under the same illumination settings, but from different camera views. Fifth and seventh columns: generated defect segmentation masks. (c) Close-ups of real (client-provided) and simulated defects.}
\end{figure}

To generate defects, we utilize 20 defective parts from the manufacturer, depicted in Figure \ref{vertical}, first column. All defective parts exhibit the very same defect typology: holes. Such open blow holes arise during the injection molding process, when gas or air is trapped in the cast. The shape and depth of these holes vary considerably among the samples: from small shallow round-shaped holes to large, deep cavities with irregular edges.  

For the experimental study, the focus is set on generating near round-shaped defects. To do so, we start with a circle as a basic shape, which we mix with different types of noises, such as fractal and turbulent noises. We randomize shape and noise parameters to generate defects of various aspects (Figure \ref{texturing_examples}, third column). During the rendering phase, we use the generated grayscale height maps of the defects as displacement maps \cite{cook1984shade} which deform polygons into a 3D mesh. Because we use bump maps \cite{blinn1978simulation}  (fake details) for part texturing  and displacement maps (true details) for defect generation, the two types of maps do not impede each other. Hence, we can vary the strength of the texture details as well as the depth of the defects independently, without needing to balance between the two. Figure \ref{texturing_examples}, from fourth to seventh columns, shows simulated images of a textured plate with corresponding defect segmentation masks. Note that the rendering was conducted under the same illumination and optical settings, but from different camera views. The rendered images reflect the real-world situations: defects visible from one viewpoint can be hardly seen from another viewpoint.

\subsection{Defect detection with deep learning}
\label{defect_detection_deep_learning}

In a supervised setting, defect detection via deep learning can be done using either segmentation (such as U-Net \cite{ronneberger2015u}) or object detection (such as YOLO \cite{redmon2016you}). Because defects can be small, we use high-resolution input images (1920 x 1200). In order to keep the real-time inference at a high frame rate, we employ custom neural networks architectures. These architectures are inspired by traditional networks like VGG, Resnet, RetinaNet\cite{lin2017focal}, etc., but their smaller backbones allow for a faster inference time on large images. 

In the following experiments, when performing domain adaptation, we use ASS\cite{wang2020alleviating}, a technique which makes use of the available labels in the target (real) domain in order to alleviate any additional semantic-level shift. ASS simultaneously adapts the domains in the feature space (via global adaptation, or GA, which is similar to DANN\cite{ganin2016domain}) and in the output space (via semantic-level adaptation, or SA). We extend the method to a fully-convolutional setup, similarly to \say{FCNs in the Wild}\cite{hoffman2016fcns}. We use gradient reversal layers \cite{ganin2016domain} instead of adversarially training the two modules, with $\lambda$ set to 0.001 for both penalties.

\section{EXPERIMENTS}
\label{sec:experiments}  

In our learning experiments on the generated simulated data, we will answer the following questions:
\begin{enumerate}[label=(\alph*)]
    \item Can we achieve good performances with the simulated images alone? 
    \item Can we improve (a) by using additional healthy real images? This is an important question since healthy parts (hence images) are easily collectable in large quantities, as opposed to defective parts, which we may not have at our disposal in case the defect is a rare event.
    \item Can we leverage simulations as additional inputs to improve over models trained with real data only? The hope is that simulations may improve robustness against defect variations and external condition changes.
    \item Can domain adaptation help the (b) and (c) models generalize better?
\end{enumerate}

\subsection{Training and evaluation data}
In total, we generated 9530 simulated images of parts from our client, which are split into a train (8576 images) and a test (954 images) set with a 9:1 ratio. Each generated simulated image contains at least one defect. We also collect real images directly from the production line of the industrial company factory. We collect 8793 images for the training set, each of which also contains at least one defect. This allows us to have comparable simulated and real training sets in terms of number of images and defects. Additionally, 8793 extra real images, all healthy, i.e. without defects, are collected. Note that the amount of extra healthy images corresponds to the number of images present in the real training set - to ensure comparability of trainings (see next section).

To evaluate the model in real life conditions, we construct a real test set of 700 images (468 defectives, 232 healthy) using different physical parts than the ones used to create the real training set. 

\subsection{Experiments and results}
\begin{table}[h!]
\centering
        \begin{tabular}{l||c|c|c|c} 
        \rule[-0.7ex]{0pt}{1.5ex}   & mAP & precision @ 0.5 & recall @ 0.5 & $f_{1}$ score @ 0.5  \\
        \hline
        \rule[-0.7ex]{0pt}{1.5ex}  Simulated	& 0.54	& 0.53	& 0.56	& 0.54 \\
        \rule[-0.7ex]{0pt}{1.5ex} \makecell[l]{Simulated + additional healthy} &	0.38 & 0.68 & 0.10 & 0.17 \\
        \rule[-0.7ex]{0pt}{1.5ex}  Real & 0.72 & 0.68 & 0.64 & 0.66 \\
        \rule[-0.7ex]{0pt}{1.5ex} Real + simulated & 0.75 & 0.69 & \textbf{0.72} & 0.70 \\
        \rule[-0.7ex]{0pt}{1.5ex} \makecell[l]{ Real + simulated w/ ASS} & \textbf{0.76} & \textbf{0.71} & 0.71 & \textbf{0.71}   \\
\end{tabular}
\caption{Performance results: mAP of each model on the real test set after 200 epochs.  Precision, recall, and $f_{1}$ score at threshold 0.5 are also shown.}
\label{tab:results}
\end{table}

In order to assess the quality of the generated simulated images, as well as to answer the four questions raised above, we train models with different training data: first with simulated images only, second with simulated and additional healthy real images, then with the real training set images only, and finally with both the real and the simulated training set images. All models are trained for 200 epochs using Adam optimizer, with a learning rate of 0.0001. We evaluate the model performance on our real test set using mean Average Precision (mAP), as well as precision, recall, and $f_{1}$ score at threshold 0.5.

\begin{figure}
    \centering
    \includegraphics[height=11cm]{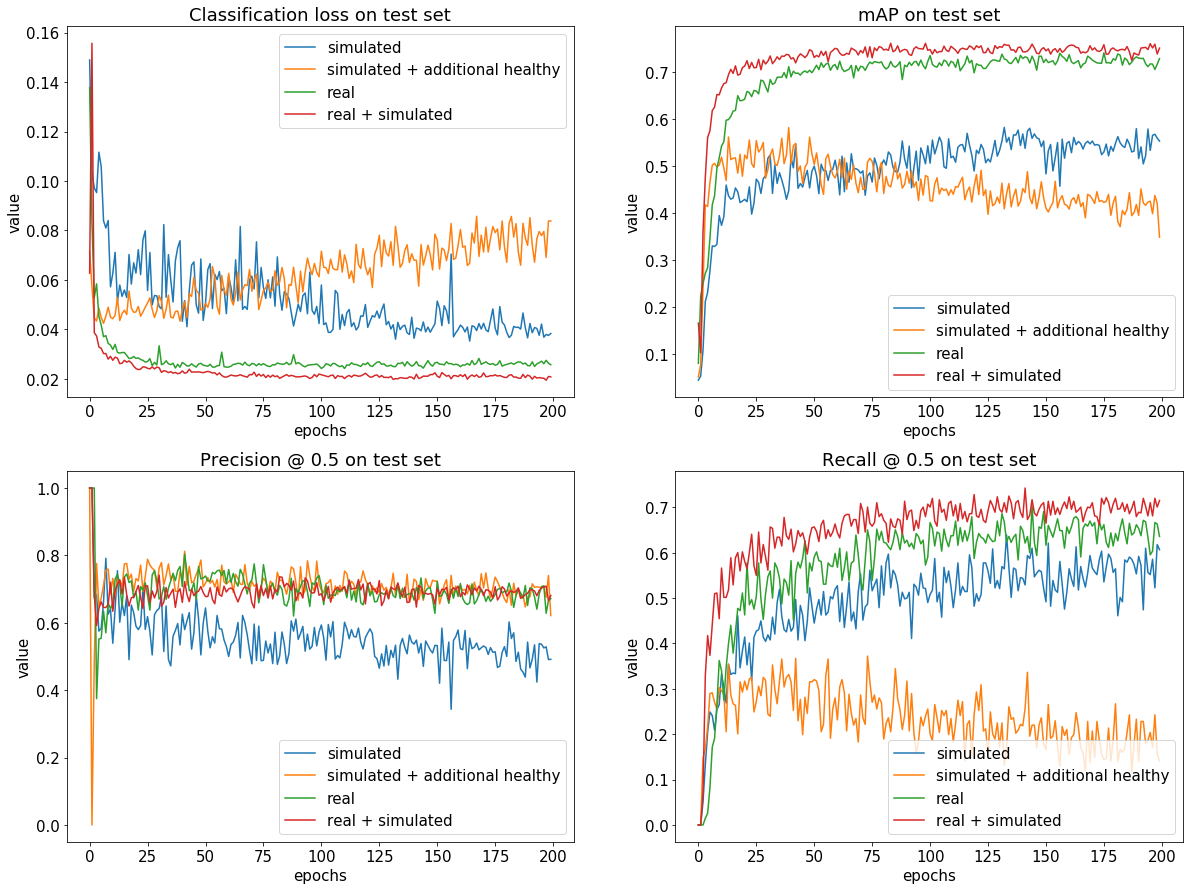}
   \caption{Classification loss, mAP, precision and recall on the real test set over training epochs.}
   \label{fig:map_loss}
\end{figure}

Figure \ref{fig:map_loss} shows the mAP and classification loss curves obtained on the real test set. Performance results for all models after 200 epochs are summarized in Table \ref{tab:results}. The following can be observed: First, with simulated data only, we can already achieve a decent mAP of 0.54 (blue curve), which has to be compared to the mAP of 0.72 which is obtained when using equivalent real data (green curve). This is encouraging, as it shows that a model trained on purely simulated data may generalize to real world data, although a large mAP performance gap remains. This answers question (a).

Second, can we improve the above training by making use of additional real healthy images on top of the simulated (defective) images? The corresponding training shows clear signs of over-fitting (orange curve): the loss on the test set clearly starts increasing after only 6 epochs, while the mAP decreases after 40 epochs. We hypothesize that this is due to the model implicitly learning to associate images coming from the real domain with “healthyness” (absence of defect). Indeed, it is easy for the model to differentiate between the two domains using low-level image statistics and textures. This information can then be applied to learn to never predict defects on the real domain - since no real defective image with defect is ever seen. The fact that we notice a strong recall drop with a stable precision strengthens this hypothesis. In order to cope with this issue, we experimented with several unsupervised and supervised domain adaptation techniques (DANN \cite{ganin2016domain}, Wasserstein DANN \cite{shen2018wasserstein}, CORAL\cite{sun2016deep}, Fast Photo Style \cite{dundar2018domain}, FDA \cite{yang2020fda}, ASS \cite{wang2020alleviating}) 
to try to make full use of the data. However, results were disappointing: in the “simulated + additional healthy” setup, no method manages to beat the simulated-only corresponding baseline (data not shown). This answers question (b).

Then, using both simulated and real training sets, we achieve an even better performance than real only, with a mAP gain of around 0.03 points (red curve). This demonstrates that simulation can be used to complement real data in order to make the model more robust, though the gain is rather small. This answers question (c).

Finally, after experimenting with all of the aforementioned domain adaptations techniques on the “real + simulated” scenario, only ASS\cite{wang2020alleviating} results in a slight mAP increment of around 0.01 point after 200 epochs (other techniques results not shown). Alongside the relatively weak performance on the “simulated + additional healthy” setup, this result suggests that domain adaptation strategies may not the best value provider. Additionally, these methods can be quite cumbersome to implement in practice. Consequently, emphasis should probably be placed on both improving photorealism of generated images and employing domain randomization in order to enhance our results. This answers question (d).

\section{Conclusions}
\label{sec:conclusion}  

In this work, we created a generic pipeline to render industrial images. Our image synthesis approach can be applied as a blueprint to many other industrial applications, even though it still requires available parametric models for each individual use case. These models can be captured using physical sensors or simulated based on statistical measurements. We demonstrate that this approach can successfully be applied to visual inspection with deep learning. We already achieve proper learning results using simulated images only (relatively to using real images only), and combining both real and simulated data results in the best performance, though there is still room for improvement. One possible option would be to increase domain randomization, which was not the focus of this paper, but is a reasonable follow-up. This could help diminish the overall engineering effort needed to achieve photorealism. In the future, we plan to further work on training with simulated defects and healthy data, as this would allow to better handle the “theoretical defect” case, in which no real image of the defect is available.

\appendix    

{\small
\bibliography{report} 

\begin{thebibliography}{10}

\bibitem{tremblay2018training}
Tremblay, J., Prakash, A., Acuna, D., Brophy, M., Jampani, V., Anil, C., To,
  T., Cameracci, E., Boochoon, S., and Birchfield, S., ``Training deep networks
  with synthetic data: Bridging the reality gap by domain randomization,'' in
  [{\em Proceedings of the IEEE Conference on Computer Vision and Pattern
  Recognition Workshops}{\nolinebreak\hspace{0.1em}]},   969--977 (2018).

\bibitem{dong2019pga}
Dong, H., Song, K., He, Y., Xu, J., Yan, Y., and Meng, Q., ``Pga-net: Pyramid
  feature fusion and global context attention network for automated surface
  defect detection,'' {\em IEEE Transactions on Industrial Informatics}~{\bf
  16}(12),  7448--7458 (2019).

\bibitem{ronneberger2015u}
Ronneberger, O., Fischer, P., and Brox, T., ``U-net: Convolutional networks for
  biomedical image segmentation,'' in [{\em International Conference on Medical
  image computing and computer-assisted
  intervention}{\nolinebreak\hspace{0.1em}]},   234--241, Springer (2015).

\bibitem{chen2017deeplab}
Chen, L.-C., Papandreou, G., Kokkinos, I., Murphy, K., and Yuille, A.~L.,
  ``Deeplab: Semantic image segmentation with deep convolutional nets, atrous
  convolution, and fully connected crfs,'' {\em IEEE transactions on pattern
  analysis and machine intelligence}~{\bf 40}(4),  834--848 (2017).

\bibitem{redmon2016you}
Redmon, J., Divvala, S., Girshick, R., and Farhadi, A., ``You only look once:
  Unified, real-time object detection,'' in [{\em Proceedings of the IEEE
  conference on computer vision and pattern
  recognition}{\nolinebreak\hspace{0.1em}]},   779--788 (2016).

\bibitem{lin2017focal}
Lin, T.-Y., Goyal, P., Girshick, R., He, K., and Doll{\'a}r, P., ``Focal loss
  for dense object detection,'' in [{\em Proceedings of the IEEE international
  conference on computer vision}{\nolinebreak\hspace{0.1em}]},   2980--2988
  (2017).

\bibitem{bergmann2019mvtec}
Bergmann, P., Fauser, M., Sattlegger, D., and Steger, C., ``Mvtec ad--a
  comprehensive real-world dataset for unsupervised anomaly detection,'' in
  [{\em Proceedings of the IEEE/CVF Conference on Computer Vision and Pattern
  Recognition}{\nolinebreak\hspace{0.1em}]},   9592--9600 (2019).

\bibitem{zambal2019end}
Zambal, S., Heindl, C., Eitzinger, C., and Scharinger, J., ``End-to-end defect
  detection in automated fiber placement based on artificially generated
  data,'' in [{\em Fourteenth International Conference on Quality Control by
  Artificial Vision}{\nolinebreak\hspace{0.1em}]},   {\bf 11172},  111721G,
  International Society for Optics and Photonics (2019).

\bibitem{li2019aads}
Li, W., Pan, C., Zhang, R., Ren, J., Ma, Y., Fang, J., Yan, F., Geng, Q.,
  Huang, X., Gong, H., et~al., ``Aads: Augmented autonomous driving simulation
  using data-driven algorithms,'' {\em Science robotics}~{\bf 4}(28) (2019).

\bibitem{tobin2017domain}
Tobin, J., Fong, R., Ray, A., Schneider, J., Zaremba, W., and Abbeel, P.,
  ``Domain randomization for transferring deep neural networks from simulation
  to the real world,'' in [{\em 2017 IEEE/RSJ international conference on
  intelligent robots and systems (IROS)}{\nolinebreak\hspace{0.1em}]},
  23--30, IEEE (2017).

\bibitem{chen2016synthesizing}
Chen, W., Wang, H., Li, Y., Su, H., Wang, Z., Tu, C., Lischinski, D., Cohen-Or,
  D., and Chen, B., ``Synthesizing training images for boosting human 3d pose
  estimation,'' in [{\em 2016 Fourth International Conference on 3D Vision
  (3DV)}{\nolinebreak\hspace{0.1em}]},   479--488, IEEE (2016).

\bibitem{sundermeyer2018learning}
Sundermeyer, M., Puang, E.~Y., Marton, Z.-C., Durner, M., and Triebel, R.,
  ``Learning implicit representations of 3d object orientations from rgb,''
  (2018).

\bibitem{langlois2019fly}
Langlois, J., Mouch{\`e}re, H., Normand, N., and Viard-Gaudin, C., ``On the fly
  generated data for industrial part orientation estimation with deep neural
  networks,'' in [{\em Fourteenth International Conference on Quality Control
  by Artificial Vision}{\nolinebreak\hspace{0.1em}]},   {\bf 11172},  111720I,
  International Society for Optics and Photonics (2019).

\bibitem{deschaintre2018single}
Deschaintre, V., Aittala, M., Durand, F., Drettakis, G., and Bousseau, A.,
  ``Single-image svbrdf capture with a rendering-aware deep network,'' {\em ACM
  Transactions on Graphics (ToG)}~{\bf 37}(4),  1--15 (2018).

\bibitem{movshovitz2016useful}
Movshovitz-Attias, Y., Kanade, T., and Sheikh, Y., ``How useful is
  photo-realistic rendering for visual learning?,'' in [{\em European
  Conference on Computer Vision}{\nolinebreak\hspace{0.1em}]},   202--217,
  Springer (2016).

\bibitem{hodavn2019photorealistic}
Hoda{\v{n}}, T., Vineet, V., Gal, R., Shalev, E., Hanzelka, J., Connell, T.,
  Urbina, P., Sinha, S.~N., and Guenter, B., ``Photorealistic image synthesis
  for object instance detection,'' in [{\em 2019 IEEE International Conference
  on Image Processing (ICIP)}{\nolinebreak\hspace{0.1em}]},   66--70, IEEE
  (2019).

\bibitem{dwibedi2017cut}
Dwibedi, D., Misra, I., and Hebert, M., ``Cut, paste and learn: Surprisingly
  easy synthesis for instance detection,'' in [{\em Proceedings of the IEEE
  International Conference on Computer Vision}{\nolinebreak\hspace{0.1em}]},
  1301--1310 (2017).

\bibitem{prakash2019structured}
Prakash, A., Boochoon, S., Brophy, M., Acuna, D., Cameracci, E., State, G.,
  Shapira, O., and Birchfield, S., ``Structured domain randomization: Bridging
  the reality gap by context-aware synthetic data,'' in [{\em 2019
  International Conference on Robotics and Automation
  (ICRA)}{\nolinebreak\hspace{0.1em}]},   7249--7255, IEEE (2019).

\bibitem{lee2019automatic}
Lee, Y.-H., Chuang, C.-C., Lai, S.-H., and Jhang, Z.-J., ``Automatic generation
  of photorealistic training data for detection of industrial components,'' in
  [{\em 2019 IEEE International Conference on Image Processing
  (ICIP)}{\nolinebreak\hspace{0.1em}]},   2751--2755, IEEE (2019).

\bibitem{akkaya2019solving}
Akkaya, I., Andrychowicz, M., Chociej, M., Litwin, M., McGrew, B., Petron, A.,
  Paino, A., Plappert, M., Powell, G., Ribas, R., et~al., ``Solving rubik's
  cube with a robot hand,'' {\em arXiv preprint arXiv:1910.07113}  (2019).

\bibitem{toldo2020unsupervised}
Toldo, M., Maracani, A., Michieli, U., and Zanuttigh, P., ``Unsupervised domain
  adaptation in semantic segmentation: a review,'' {\em Technologies}~{\bf
  8}(2),  35 (2020).

\bibitem{wang2020alleviating}
Wang, Z., Wei, Y., Feris, R., Xiong, J., Hwu, W.-M., Huang, T.~S., and Shi, H.,
  ``Alleviating semantic-level shift: A semi-supervised domain adaptation
  method for semantic segmentation,'' in [{\em Proceedings of the IEEE/CVF
  Conference on Computer Vision and Pattern Recognition
  Workshops}{\nolinebreak\hspace{0.1em}]},   936--937 (2020).

\bibitem{goodfellow2014generative}
Goodfellow, I.~J., Pouget-Abadie, J., Mirza, M., Xu, B., Warde-Farley, D.,
  Ozair, S., Courville, A., and Bengio, Y., ``Generative adversarial
  networks,'' {\em arXiv preprint arXiv:1406.2661}  (2014).

\bibitem{hoffman2018cycada}
Hoffman, J., Tzeng, E., Park, T., Zhu, J.-Y., Isola, P., Saenko, K., Efros, A.,
  and Darrell, T., ``Cycada: Cycle-consistent adversarial domain adaptation,''
  in [{\em International conference on machine
  learning}{\nolinebreak\hspace{0.1em}]},   1989--1998, PMLR (2018).

\bibitem{dundar2018domain}
Dundar, A., Liu, M.-Y., Wang, T.-C., Zedlewski, J., and Kautz, J., ``Domain
  stylization: A strong, simple baseline for synthetic to real image domain
  adaptation,'' {\em arXiv preprint arXiv:1807.09384}  (2018).

\bibitem{yang2020fda}
Yang, Y. and Soatto, S., ``Fda: Fourier domain adaptation for semantic
  segmentation,'' in [{\em Proceedings of the IEEE/CVF Conference on Computer
  Vision and Pattern Recognition}{\nolinebreak\hspace{0.1em}]},   4085--4095
  (2020).

\bibitem{ganin2016domain}
Ganin, Y., Ustinova, E., Ajakan, H., Germain, P., Larochelle, H., Laviolette,
  F., Marchand, M., and Lempitsky, V., ``Domain-adversarial training of neural
  networks,'' {\em The Journal of Machine Learning Research}~{\bf 17}(1),
  2096--2030 (2016).

\bibitem{hoffman2016fcns}
Hoffman, J., Wang, D., Yu, F., and Darrell, T., ``Fcns in the wild: Pixel-level
  adversarial and constraint-based adaptation,'' {\em arXiv preprint
  arXiv:1612.02649}  (2016).

\bibitem{shen2018wasserstein}
Shen, J., Qu, Y., Zhang, W., and Yu, Y., ``Wasserstein distance guided
  representation learning for domain adaptation,'' in [{\em Proceedings of the
  AAAI Conference on Artificial Intelligence}{\nolinebreak\hspace{0.1em}]},
  {\bf 32}(1) (2018).

\bibitem{sun2016deep}
Sun, B. and Saenko, K., ``Deep coral: Correlation alignment for deep domain
  adaptation,'' in [{\em European conference on computer
  vision}{\nolinebreak\hspace{0.1em}]},   443--450, Springer (2016).

\bibitem{vu2019advent}
Vu, T.-H., Jain, H., Bucher, M., Cord, M., and P{\'e}rez, P., ``Advent:
  Adversarial entropy minimization for domain adaptation in semantic
  segmentation,'' in [{\em Proceedings of the IEEE/CVF Conference on Computer
  Vision and Pattern Recognition}{\nolinebreak\hspace{0.1em}]},   2517--2526
  (2019).

\bibitem{dahmen2019digital}
Dahmen, T., Trampert, P., Boughorbel, F., Sprenger, J., Klusch, M., Fischer,
  K., Kübel, C., and Slusallek, P., ``Digital reality: a model-based approach
  to supervised learning from synthetic data,'' {\em AI Perspectives}~{\bf 1},
  1--12 (2019).

\bibitem{remondino2005photogrammetry}
Remondino, F., Guarnieri, A., and Vettore, A., ``3d modeling of close-range
  objects: photogrammetry or laser scanning?,'' in [{\em Proceedings of
  SPIE-IS\&T Electronic Imaging: Videometrics \MakeUppercase{\romannumeral
  8}}{\nolinebreak\hspace{0.1em}]},   {\bf 5665},  216--225 (2004).

\bibitem{woodham1980photometric}
Woodham, R.~J., ``Photometric method for determining surface orientation from
  multiple images,'' {\em Optical Engineering}~{\bf 19}(1) (1980).

\bibitem{blinn1978simulation}
Blinn, J.~F., ``Simulation of wrinkled surfaces,'' in [{\em Proceedings of
  SIGGRAPH 78}{\nolinebreak\hspace{0.1em}]},   286--292 (1978).

\bibitem{criminisi2004inpainting}
Criminisi, A., P\'erez, P., and Toyama, K., ``Region filling and object removal
  by exemplar-based image inpainting,'' {\em IEEE Transactions on Image
  Processing}~{\bf 13}(9),  1200--1212 (2004).

\bibitem{trampert2018inpainting}
Trampert, P., Schlabach, S., Dahmen, T., and Slusallek, P., ``Exemplar-based
  inpainting based on dictionary learning for sparse scanning electron
  microscopy,'' {\em Microscopy and Microanalysis}~{\bf 24}(S1),  700--701
  (2018).

\bibitem{cook1984shade}
Cook, R.~L., ``Shade trees,'' in [{\em Proceedings of SIGGRAPH
  84}{\nolinebreak\hspace{0.1em}]},   {\bf 18}(3),  223--231 (1984).

\end{thebibliography}
\bibliographystyle{spiebib} 
}
\end{document}